\documentclass[acmsmall, review=False, screen=true]{acmart}

\setcopyright{acmcopyright}
\acmJournal{TKDD}
\acmYear{2024} 
\acmVolume{1} \acmNumber{1} \acmArticle{1} \acmMonth{1} \acmPrice{15.00}
\acmDOI{00.0000/0000000}

\AtBeginDocument{%
  }

\usepackage{hyperref}
\usepackage{url}
\usepackage{graphicx}
\usepackage{xspace}
\usepackage{booktabs}
\usepackage{caption, subcaption}
\usepackage{multirow}
\usepackage{amsmath}
\usepackage{amsfonts} 
\usepackage{hyperref}
\usepackage{multicol}
\usepackage{pifont}
\usepackage{tikz}

\usepackage[usestackEOL]{stackengine}

\newcommand{\icewsw}{ICEWS18\xspace}
\newcommand{\icewsd}{ICEWS14\xspace}

\newcommand{\calM}{\mathcal{M}}

\newcommand{\tr}{D^{train}}
\newcommand{\te}{D^{test}}
\newcommand{\val}{D^{val}}

\setcopyright{acmlicensed}
\copyrightyear{2024}
\acmYear{2024}
\acmDOI{XXXXXXX.XXXXXXX}

\acmConference[Conference acronym 'XX]{Make sure to enter the correct
  conference title from your rights confirmation email}{June 03--05,
  2018}{Woodstock, NY}
\acmISBN{978-1-4503-XXXX-X/18/06}

\begin{document}
\emergencystretch 2em

\title{Towards Improving Long-Tail Entity Predictions in Temporal Knowledge Graphs through Global Similarity and Weighted Sampling}

\author{Mehrnoosh Mirtaheri}
\affiliation{%
  \institution{USC Information Sciences Institute}
  \city{Los Angeles}
  \state{CA}
    \country{USA}}

\author{Ryan A. Rossi}
\affiliation{%
  \institution{Adobe Research}
  \city{San Jose}
  \state{CA}
    \country{USA}
 }

\author{Sungchul Kim}
\affiliation{%
  \institution{Adobe Research}
  \city{San Jose}
  \state{CA}
    \country{USA}}

\author{Kanak Mahadik}
\affiliation{%
  \institution{Adobe Research}
  \city{San Jose}
  \state{CA}
    \country{USA}}

\author{Tong Yu}
\affiliation{%
  \institution{Adobe Research}
  \city{San Jose}
  \state{CA}
    \country{USA}}

\author{Xiang Chen}
\affiliation{%
  \institution{Adobe Research}
  \city{San Jose}
  \state{CA}
    \country{USA}}

\author{Mohammad Rostami}
\affiliation{%
\institution{USC Information Sciences Institute}
\city{Los Angeles}
  \state{CA}
    \country{USA}}

\renewcommand{\shortauthors}{Mirtaheri et al.}

\begin{abstract}
Temporal Knowledge Graph (TKG) completion models traditionally assume access to the entire graph during training. This overlooks challenges stemming from the evolving nature of TKGs, such as: (i) the model's requirement to generalize and assimilate new knowledge, and (ii) the task of managing new or unseen entities that often have sparse connections. In this paper, we present an incremental training framework specifically designed for TKGs, aiming to address entities that are either not observed during training or have sparse connections. Our approach combines a model-agnostic enhancement layer with a weighted sampling strategy, that can be augmented to and improve any existing TKG completion method. The enhancement layer leverages a broader, global definition of entity similarity, which moves beyond mere local neighborhood proximity of GNN-based methods. The weighted sampling strategy employed in training accentuates edges linked to infrequently occurring entities. We evaluate our method  on two benchmark datasets, and demonstrate that our framework outperforms existing methods in total link prediction, inductive link prediction, and in addressing long-tail entities. Notably, our method achieves a 10\% improvement and a 15\% boost in MRR for these datasets. The results underscore the potential of our approach in mitigating catastrophic forgetting and enhancing the robustness of TKG completion methods, especially in an incremental training context \footnote{Preliminary results of this work are presented in the 2024 ACM on Web Conference \citep{mirtaheri2024tackling}.}. \end{abstract}

 %




\maketitle

\section{Introduction}

Knowledge graphs (KGs) have become foundational tools for structuring and representing multi-relational data, playing a critical role in advancing a wide range of natural language processing (NLP) tasks, such as question answering, information retrieval, and reasoning about complex queries using large language models (LLMs) \citep{pan2023unifying}. At their core, KGs encode factual knowledge using a triplet-based schema—comprising subject, relation, and object—that captures relationships between entities in a structured and interpretable manner \citep{liang2022reasoning}. To capture the temporal dynamics of real-world knowledge, temporal knowledge graphs (TKGs) extend this representation by incorporating time as an additional dimension. This can be broadly categorized into two primary types: semantic KGs, which encode facts valid over time intervals (e.g., roles or attributes that persist), and event-centric TKGs, which capture discrete, timestamped events emphasizing the granularity and recurrence of interactions. 
Semantic KGs, such as YAGO \citep{kasneci2009yago}, encode facts that persist over specific time intervals—e.g., \textit{(Obama, President, United States, 2009–2017)}—allowing for reasoning over temporal scopes of entity roles and attributes. In contrast, event-centric TKGs, such as ICEWS \citep{boschee2015integrated}, capture discrete events annotated with precise timestamps—e.g., \textit{(Obama, meet, Merkel)} at distinct dates—emphasizing temporal granularity and the recurrence of interactions.
By modeling both enduring relationships and transient events, TKGs offer a more comprehensive framework for understanding how entities and their interactions evolve over time. This temporal dimension is particularly valuable for downstream applications that require temporal reasoning, trend analysis, or historical inference, thereby extending the utility of KGs beyond static knowledge representation.

\begin{figure*}
    \centering
    \begin{subfigure}[b]{0.35\textwidth}
        \centering
        \includegraphics[width=\textwidth]{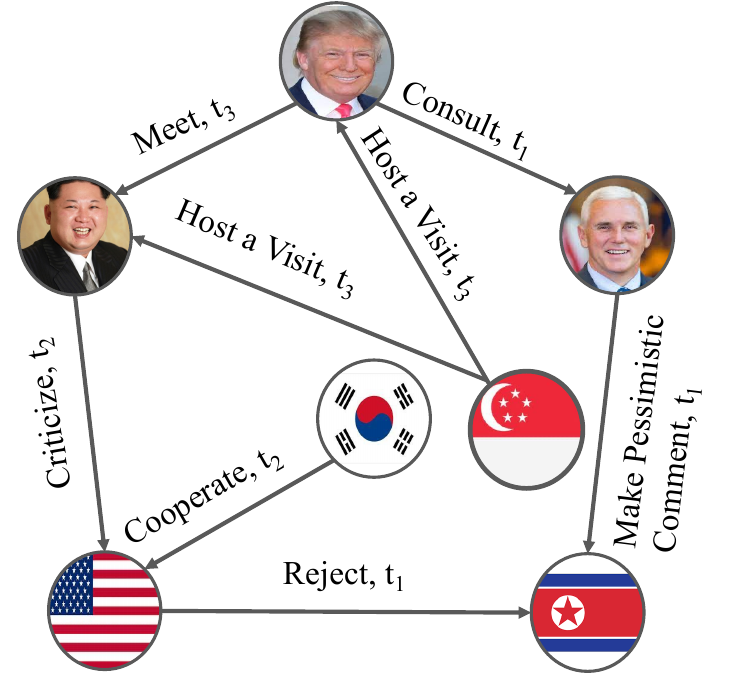}
        \vspace{1pt}
        \caption{Temporal Knowledge Graph.} 
        \label{fig:tkg-graph}
    \end{subfigure}
    \hfill
    \begin{subfigure}[b]{0.64\textwidth}
        \centering
        \includegraphics[width=\textwidth]{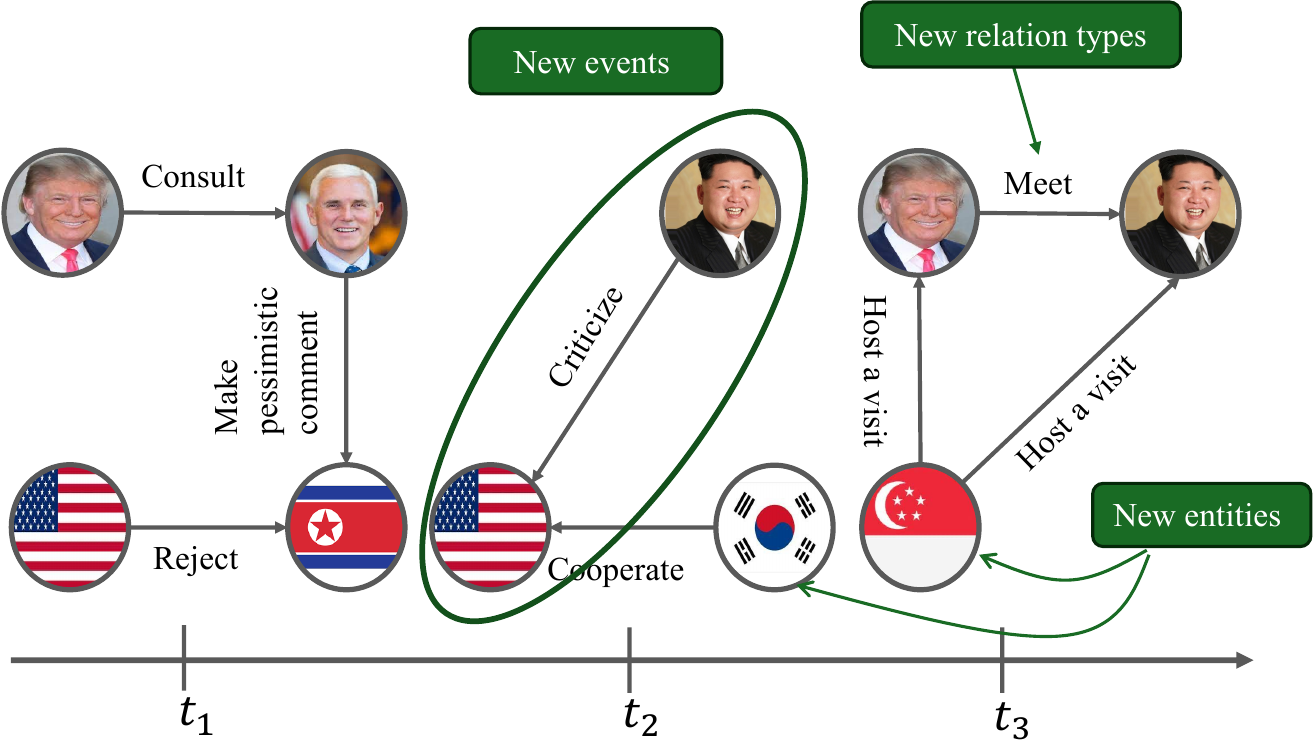}
        \caption{Incremental updates in the graph.} 
        \label{fig:tkg-graph-incremental}
    \end{subfigure}  
    \caption{Illustration of the Temporal Knowledge Graph and its incremental updates.}
    \label{fig:tkg-overview}
\end{figure*}

A significant limitation of both traditional knowledge graphs (KGs) and temporal knowledge graphs (TKGs) is their inherent incompleteness, which hampers the effectiveness of downstream tasks such as link prediction, reasoning, and entity alignment. While advances in automatic knowledge extraction have led to improved KG coverage, the issue of missing or incomplete facts remains pervasive and problematic. This limitation is especially pronounced in real-world KGs derived from dynamic and continuously evolving data sources, such as live news streams, social media, or sensor networks. In these scenarios, new facts are generated in real time, often accompanied by a temporal context and introduce previously unseen entities and relations, thereby compounding the challenge of maintaining an up-to-date and coherent KG (Figure \ref{fig:tkg-overview}).

To address these challenges, a large body of research has focused on knowledge graph completion (KGC) \citep{wang2022simkgc, huang2022multilingual, shen2022comprehensive}, with particular emphasis on temporal KGC methods that model time-aware representations by leveraging historical interaction patterns \citep{jin2019recurrent, zhu2020learning, sun2021timetraveler}. These approaches typically encode temporal dependencies and trends to infer plausible missing links across different time steps. However, a major limitation of many existing methods is the unrealistic assumption of full graph observability during training. In practice, this assumption fails to accommodate two key challenges inherent to real-world TKGs: (i) handling shifts in the entity distribution, such as the frequent emergence of new entities or the sporadic reappearance of previously seen ones, and (ii) enabling continual or incremental integration of new facts without catastrophic forgetting of previously acquired knowledge.
Addressing these issues is crucial for the development of robust and scalable TKG systems capable of operating in non-stationary, streaming environments. Effective solutions must strike a balance between adaptability to new information and retention of long-term knowledge, calling for methods that support dynamic updates, continual learning, and generalization to previously unseen entities and relations.

Conventional TKG completion methods predominantly rely on static embedding tables, where each entity observed during training is assigned a learned vector representation \citep{jin2019recurrent, wu2020temp}. This design introduces an important limitation: entities that are unseen during training cannot be represented, as no pre-initialized embedding exists for them. Consequently, models built on this architecture are unable to make predictions involving such entities at inference time—a   drawback in real-world applications, where new entities continuously emerge in evolving temporal data streams.
To mitigate this issue, some approaches have incorporated Graph Neural Network (GNN) architectures that aim to dynamically infer embeddings for unseen entities by aggregating information from their temporal neighborhoods. However, the effectiveness of these methods is often constrained by the sparsity and limited connectivity of new entities, which restricts the amount of contextual information available for embedding construction. As a result, inferred embeddings for such entities tend to be noisy or insufficiently informative, leading to suboptimal performance in downstream tasks.
Despite the importance of this challenge, only a handful of studies explicitly evaluate their models under an inductive link prediction setting—where test-time queries involve entities that were not present during training \citep{han2020xerte, sun2021timetraveler}. Even fewer propose architectural innovations specifically designed to enhance representation learning and predictive accuracy for these previously unseen entities. The lack of focus on this inductive generalization capability leaves a critical gap in the development of truly robust and adaptable TKG models capable of operating in open-world environments.

The challenges  extend beyond the scope of representation learning to encompass the broader problem of model adaptation in the presence of continuously evolving knowledge. While certain methods attempt to address inductive settings by generating embeddings for newly encountered entities at test time \citep{han2020xerte}, they often fall short in one crucial aspect: adapting the model’s internal parameters to reflect the semantics and relational context introduced by these new entities and their associated facts. In essence, the underlying predictive mechanisms of the model remain static, even as the graph structure evolves dynamically.
A straightforward solution—retraining the model from scratch to incorporate newly acquired facts—quickly becomes impractical in real-world scenarios due to the immense computational cost and latency involved, particularly for large-scale TKGs. On the other hand, naive fine-tuning approaches risk overfitting to the new data and suffer from catastrophic forgetting, wherein previously learned patterns and temporal dependencies are overwritten or degraded. To mitigate this, recent work has proposed incremental training frameworks for TKG completion that aim to retain prior knowledge while integrating new information in a more efficient and stable manner \citep{wu2021tie, mirtaheri2023history}. These methods typically build on a pre-trained base model, applying regularization strategies or memory replay mechanisms to avoid forgetting.
However, a critical limitation of these frameworks is that they are model-agnostic and preserve the architecture of the underlying TKG model. This presents a problem because most existing TKG models are not inherently inductive and lack the design principles needed to cope with sparse or previously unseen entities. Without specific architectural components to support dynamic generalization and contextualization for such entities, these incremental frameworks offer limited effectiveness in real-world settings where the knowledge graph is both incomplete and non-stationary. Addressing this gap calls for new model architectures or training strategies that jointly support inductive reasoning, continual adaptation, and robustness to data sparsity.

To address these challenges, we propose a solution that integrates both the incremental learning requirements and the handling of sparse or unseen entities in TKGs. Our approach consists of an incremental training framework for TKG forecasting, complemented by novel architectural components to handle entity sparsity. The core of our technical contribution is a model-agnostic enhancement module that can be integrated with existing GNN-based TKG completion methods, and is designed to enhance entity representations when local connectivity is limited. This module operates by computing entity similarities using global graph information rather than relying solely on immediate neighborhoods \cite{mirtaheri2024tackling}. Further, our framework also incorporates a frequency-based sampling mechanism that gives greater weight to training examples involving entities with limited occurrence, thereby improving the model's ability to handle both sparsely connected and newly introduced entities. To evaluate our approach, we develop and release two new benchmark datasets constructed specifically for evaluating incremental learning scenarios in TKGs. A summary of the main contributions of our work are as follows:

\begin{itemize}
    \item \textbf{Incremental Training Framework:} We propose a novel framework for incremental training in TKGs, allowing for the seamless integration of newly introduced entities while preserving existing knowledge.
    
    \item \textbf{Model-Agnostic Enhancement Layer} that can be augmented to most GNN-based TKG completion models, and improves entity representations by utilizing a broader similarity perspective rather than relying solely on local connections.
    
    \item \textbf{Weighted Sampling Strategy:} We implement a training mechanism that prioritizes edges involving infrequent entities, facilitating improved prediction for long-tail and previously unseen entities.
    
    \item \textbf{Benchmark Datasets for Incremental Learning:} To assess our framework, we curate two benchmark datasets specifically designed for evaluating incremental learning performance in TKG completion.
    
    \item \textbf{Comprehensive Evaluation:} We conduct extensive experiments, analyzing overall link prediction performance, inductive link prediction accuracy, and the effectiveness of our approach in handling entities with sparse connections.
\end{itemize}

By addressing the challenges associated with incremental learning in TKGs, our work contributes to improving the adaptability and predictive performance of knowledge graph completion methods, particularly for dynamic real-world applications.

\section{Related Work} \label{sec:related-work}
This study intersects with research on TKG completion, continual learning techniques, and the recent advancements in continual learning specifically tailored for knowledge graphs.

\textbf{Temporal Knowledge Graph Completion}. TKG completion  methods can be broadly categorized into two main paradigms based on how they incorporate temporal information: translation-based models and neural sequence-based models.
Translation-based approaches extend static knowledge graph embedding models by encoding temporal information directly into the entity and relation embeddings. These methods often introduce time-aware transformations or mappings to adjust embeddings based on timestamps. For instance, some works~\citep{leblay2018deriving,jain2020temporal} project timestamps into vector spaces and integrate them into scoring functions, whereas others~\citep{dasgupta2018hyte,wang2019hybrid} use timestamp-specific hyperplanes to alter the representation space, enabling the embeddings to shift over time. Such methods rely on a predefined temporal transition function that modulates the base embeddings of entities and relations, allowing the model to reflect the temporal evolution of facts.
Neural sequence-based approaches, on the other hand, model temporal dynamics by capturing sequential dependencies in the data. These include shallow encoders~\citep{xu2019temporal} and deeper architectures such as RNNs, GNNs, and attention-based models. DyERNIE~\citep{han2020dyernie}, for example, employs hyperbolic embeddings within a temporal sequence encoder to capture hierarchical temporal dependencies across events. Other methods, like Know-Evolve~\citep{trivedi2017know}, leverage temporal point processes to model the fine-grained dynamics of entity interactions over time. Recurrent models~\citep{jin2020recurrent,wu2020temp} aggregate historical neighborhood information to enhance reasoning over evolving graphs, learning time-sensitive relational patterns from entity histories.

While both paradigms have advanced temporal reasoning, most existing methods  assume that the set of entities and relations in the TKG remains fixed during inference. Recent studies~\citep{han2020xerte,sun2021timetraveler,han-etal-2021-learning-neural} attempt to address inductive temporal link prediction, enabling models to generalize to unseen temporal links. However, these models still fall short in supporting continual learning: the ability to incrementally update knowledge and adapt to newly emerging entities and relations over time without retraining from scratch. Thus, handling dynamically evolving TKGs with open-world assumptions remains a largely underexplored challenge.

\textbf{Continual Learning}.   Continual learning (CL), also referred to as lifelong learning, is a   learning paradigm in which a model is exposed to a sequence of tasks or data distributions that are encountered one at a time and must learn them incrementally over time. Unlike traditional batch learning, once a task is learned, its training data is no longer accessible in subsequent stages, making the learning process non-stationary and memory-constrained. A central challenge in this setting is catastrophic forgetting, where learning new tasks leads to the erosion of knowledge gained from previously learned tasks. Catastrophic forgetting occurs because updates to the model's parameters for the current task may overwrite the representations crucial for performing past tasks well, thereby degrading performance on earlier objectives.

To mitigate forgetting, a widely used approach is experience    replay~\citep{li2018learning,rostami2021lifelong}, which draws inspiration from memory consolidation mechanisms observed in biological systems. In this strategy, a limited subset of training examples from previous tasks is stored in a memory buffer. These samples are selected to represent the corrsponding task well. During training on new tasks, the model is periodically exposed to these stored samples, enabling it to reinforce older knowledge while acquiring new skills. This joint rehearsal helps stabilize the model's internal representations and maintain performance across tasks.
A key constraint of experience replay is the fixed size of the memory buffer, as unbounded memory growth is impractical in real-world applications. This limitation necessitates intelligent sample selection and retention strategies to maximize the utility of the buffer. One  proposed approach~\citep{schaul2015prioritized} is used prioritized sampling, where examples are selected based on their contribution to the training loss—favoring those that are most informative or error-prone. Other strategies include diversity-based selection, class balancing, and uncertainty-aware retention policies, all aimed at preserving a compact yet representative snapshot of the past experiences.

To eliminate the need for a memory buffer, generative models   synthesize pseudo-samples that mimic past task data, thereby preserving learned knowledge without explicitly storing task  sample~\citet{shin2017continual,rostami2019complementary}. For instance, Shin et al.~\citet{shin2017continual} employ a generative adversarial network (GAN) to reconstruct pseudo-data from previous tasks which are similar to real data, enabling rehearsal-based training while avoiding memory overhead. Alternatively, weight consolidation methods, such as Elastic Weight Consolidation (EWC)~\citep{kirkpatrick2017overcoming} and synaptic intelligence~\citep{zenke2017temporal}, mitigate catastrophic forgetting by identifying and rigidly preserving the most critical weights for past tasks. Weight consolidation ensures that new task learning primarily adapts the remaining, less constrained parameters, and benefit from knowledge transfer through consolidated weights. Our proposed framework integrates both generative replay and weight consolidation, leveraging their complementary strengths to enhance continual learning performance. Weight consolidation and experience replay are often used exclusively, but it is possible to use them together to benefit from the strengths of both approaches.

\textbf{Continual Graph Learning}. Continual learning in the context of graph-structured data remains a relatively underexplored area, particularly when compared to its extensive study in domains such as computer vision and natural language processing. Only a limited number of works have investigated continual or incremental learning on dynamic heterogeneous networks \citep{Tang2020, wang2020streaming, zhou2021overcoming} and semantic knowledge graphs (KGs) \citep{Song2018, Daruna2021, Wu2021}. Among these, early efforts such as \citep{Song2018, Daruna2021} adapt class-incremental learning paradigms to static KGs by integrating traditional translation-based embedding models like TransE \citep{bordes2013translating}. These methods attempt to incrementally update entity and relation embeddings as new classes or facts are introduced. However, they typically assume that the underlying graph structure evolves slowly or remains largely static—an assumption that fails to hold in real-world temporal KGs, where new entities, relations, and interactions emerge continuously and unpredictably.

To accommodate evolving semantics over time, the Temporal Incremental Embedding (TIE) framework \citep{Wu2021} extends continual learning to semantic KGs by discretizing time into coarse intervals (e.g., years) and decomposing each temporally extended fact into a sequence of timestamped atomic events. While this approach allows temporal reasoning across time snapshots, the coarse-grained temporal resolution can obscure fine-grained event dynamics that are often critical in real-world, event-centric KGs—such as those derived from news or transaction logs—where precise temporal relationships and orderings play a key role in inference.

More recently, \citet{mirtaheri2023history} introduced a continual TKG completion framework that combines regularization techniques with experience replay to preserve previously acquired knowledge while integrating new temporal facts. This framework effectively reduces catastrophic forgetting and demonstrates promising performance on standard benchmarks. However, a major limitation remains: the model does not explicitly account for long-tail entities—those that appear infrequently or only within specific temporal contexts. These entities are abundant in real-world TKGs and pose substantial challenges for representation learning due to their sparse interactions and limited historical context. As a result, even models with strong continual learning capabilities may struggle to generalize effectively to the tail of the distribution, leading to reduced robustness and incomplete knowledge coverage.

\section{Problem Formulation}
In this section, we define temporal knowledge graph completion and extend this foundation to our incremental training framework.

\subsection{Temporal Knowledge Graph Completion}
A temporal knowledge graph (TKG), denoted as $G = \langle \mathcal{Q}, \mathcal{E}, \mathcal{R} \rangle$, is a data structure designed to capture the evolution of relationships and facts over time. In a TKG, information is represented as a set of quadruples $ \mathcal{Q} = {(s, r, o, \tau) \mid s, o \in \mathcal{E}, r \in \mathcal{R}, \tau \in \mathbb{T} }$, where  $s$ (subject) and $o$ (object) are entities from the set of all entities $\mathcal{E}$,  $r$ is a relation type from the set of all relations $\mathcal{R}$, and   $\tau$ is a timestamp indicating when the interaction occurs.
Each quadruple $(s, r, o, \tau)$ thus encodes a fact that the relation $r$ holds between entities $s$ and $o$ at time $\tau$. For example, the quadruple (``Barack Obama'', ``presidentOf'', ``USA'', 2009) indicates that Barack Obama became president of the USA in 2009.
The timestamp length would depend on the application and use cases. Knowledge graph completion for TKGs involves inferring missing or future interactions between entities, given the observed data. Typically, this task is formulated as an entity prediction problem: given a subject $s$, a relation $r$, and a timestamp $\tau$, the goal is to predict the most likely object $o$ that completes the quadruple $(s, r, o, \tau)$. Conversely, one can also predict the subject given the object, relation, and timestamp.
For a TKG with observations up to time $T$, prediction tasks generally fall into two categories:

\begin{itemize}
    \item \textbf{Interpolation}: Reconstructing unobserved interactions that occurred within the temporal boundary of the training data (before time $T$)
    \item \textbf{Extrapolation} (or forecasting): Predicting future interactions that will occur beyond the temporal boundary of the training data (after time $T$)
\end{itemize}

By leveraging the temporal dimension, TKG completion models can capture the dynamics of evolving relationships and provide more accurate and context-aware predictions.
 focuses on the extrapolation paradigm, which presents additional challenges due to the need to generalize temporal patterns into unseen future states. 

\subsection{Incremental Learning for Temporal Knowledge Graphs}
\label{secs:cl_setup}
Real-world temporal knowledge graphs continually evolve as new data becomes available. To model this evolution, we represent a TKG $G$ as a sequence of graph snapshots $\langle G_1, G_2, \dots, G_T \rangle$ arriving over time. Each snapshot $G_t = \langle \mathcal{Q}_t, \mathcal{E}_t, \mathcal{R}_t \rangle$ contains:

\begin{itemize}
    \item A set of quadruples $\mathcal{Q}_t = \{(s, r, o, \tau) | s, o \in \mathcal{E}_t, r \in \mathcal{R}_t, \tau \in \left[ \tau_t, \tau_{t+1} \right)\}$ that occurred during the time interval $\left[\tau_t, \tau_{t+1}\right)$
    \item The corresponding entity set $\mathcal{E}_t$ and relation set $\mathcal{R}_t$ for that time period
\end{itemize}

\begin{figure*}
    \centering
    \includegraphics[width=0.9\textwidth]{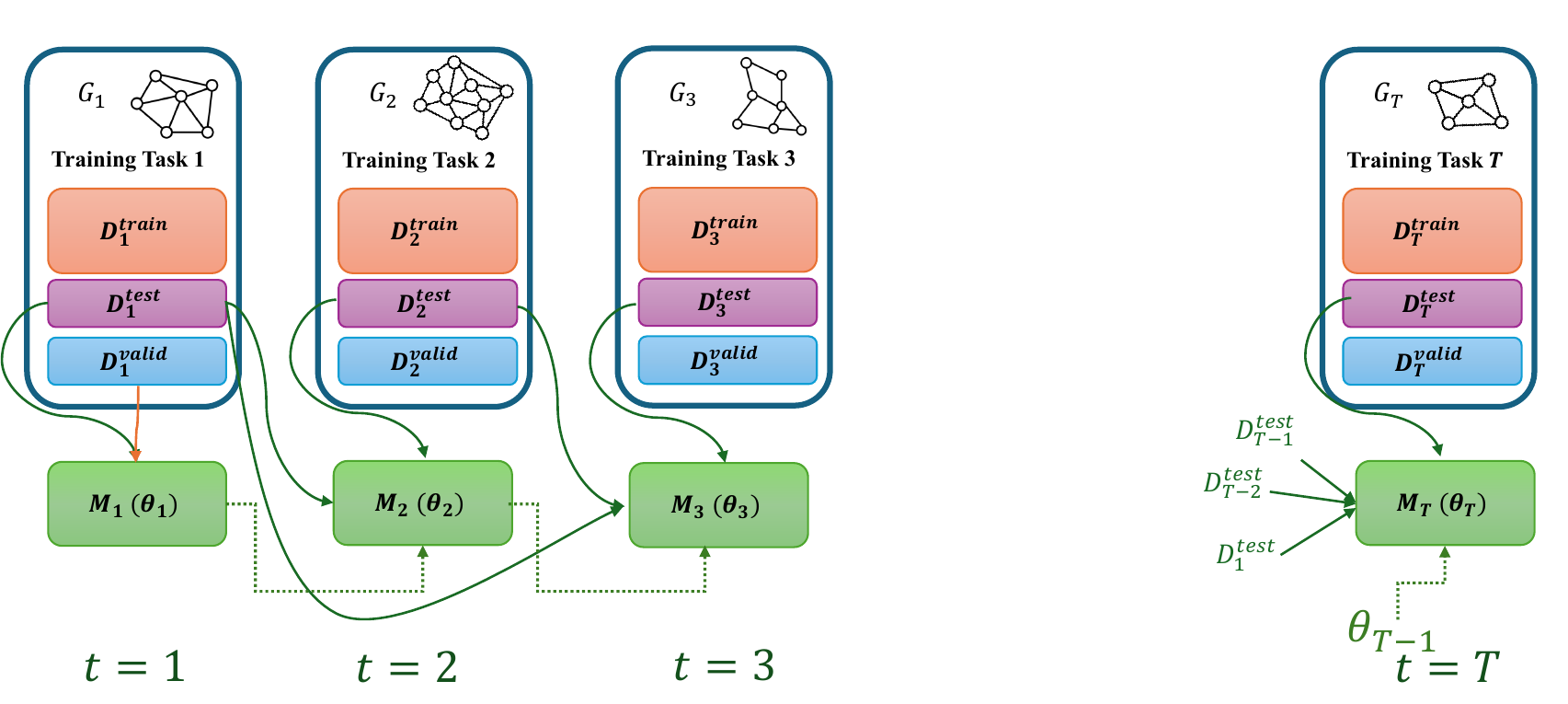}
    \includegraphics[width=0.9\textwidth]{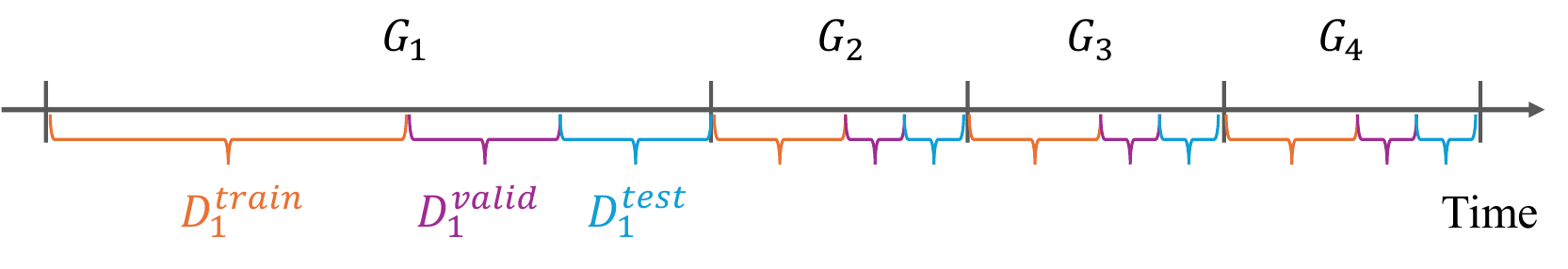}
    \caption{Continual Extrapolation Setup. The arrow depicts the time. For extrapolation, the edge timestamps in the validation and test set do not have any overlap.}
    \label{fig:cl-ext}
\end{figure*}

Since we observe these snapshots incrementally, a model trained only on the first snapshot would struggle to generalize to subsequent snapshots that might contain new entities, relations, or interaction patterns. Incremental training involves updating model parameters as new graph snapshots become available, while retaining knowledge from previous snapshots. We formalize this as a sequence of learning tasks $\langle\mathcal{T}_1, \dots, \mathcal{T}_T \rangle$, where each task $\mathcal{T}_t = \left(D_t^{train}, D_t^{test}, D_t^{val}\right)$ contains training, validation, and test data derived from snapshot $G_t$.The train, validation and test sets for each task are configured for temporal extrapolation with non-overlapping time intervals, as illustrated in Figure \ref{fig:cl-ext} (bottom).

The training process yields a sequence of models $\calM = \langle\calM_1, \dots, \calM_T\rangle$ with corresponding parameter sets $\theta = \langle \theta_1, \theta_2, ..., \theta_T\rangle$. Each model $\calM_t$ builds upon the previous model $\calM_{t-1}$ by incorporating new information while aiming to preserve previously learned patterns. Figure \ref{fig:cl-ext} illustrates our incremental training and evaluation setup for temporal extrapolation.

\section{Methodology} \label{sec:approach}
TKGs are inherently dynamic structures where entities and their relationships evolve over time. One of the fundamental challenges in TKG completion tasks is the long-tailed distribution of entities—many entities appear infrequently, leading to sparse and uneven connectivity patterns. This issue is further exacerbated in evolving TKGs, where the temporal aspect introduces additional complexity. As entities shift in relevance or interact with different subsets of other entities over time, their structural roles in the graph may fluctuate. For instance, an entity might have a high degree of connectivity during one time interval but become marginal or inactive in others.
This temporal variability gives rise to two critical challenges in modeling TKGs effectively:
\begin{enumerate}
    \item  Bias in model learning: Parameters tend to be overfitted to entities that occur frequently in recent time slices, reducing the model’s generalizability and performance on underrepresented entities.
    \item  Representation staleness: Embeddings for long-tailed or infrequent entities often become outdated as the model struggles to adapt to the evolving context in which these entities appear, resulting in inaccurate or irrelevant representations.
\end{enumerate}
To address these challenges, we propose a robust and flexible framework composed of two core components:
\begin{enumerate}
    \item  A Model-Agnostic Enhancement Layer, designed to refine entity representations and mitigate bias without modifying the underlying architecture.
    \item  A Weighted Frequency-Based Sampling strategy, aimed at dynamically adjusting the training distribution to better reflect temporal relevance and long-tail importance.
\end{enumerate}
 
In the following subsections, we detail the design and functionality of each component and demonstrate how they collectively improve model robustness and long-tail generalization in dynamic TKG settings.

\subsection{Model-Agnostic Enhancement Layer}
The model-agnostic enhancement layer extends beyond the typical local neighborhood constraints in existing GNN-based knowledge graph completion models. Instead of limiting information flow to an entity's immediate connections, this approach augments entity embeddings by incorporating relevant signals from semantically similar entities across the entire graph, regardless of their topological distance. When integrated with TKG completion frameworks, this enhancement mechanism addresses the fundamental challenge of sparse connectivity patterns. The enhancement layer constructs enriched entity representations through the following formulation:

\begin{equation}
\label{eq:enhancement_layer}
e_s = \lambda f(s) + \phi(d_s)(1-\lambda) g(s)
\end{equation}

This equation combines two components: $f(s)$, which captures the base entity representation from the underlying model architecture, and $g(s)$, a temporal enhancement function that aggregates information from similar entities $S_t(s)$ to $s$ at time $t$. The term $\phi (d_s)$ is a decreasing function of the degree of entity $s$. Intuitively, entities with fewer connections receive stronger enhancement effects. The $\lambda$ parameter controls the balance between original and enhanced representations.  This enhancement framework provides flexibility in choosing the similarity criteria and the definition and enhancement function ($g(s)$) which can be changed depending on the application.\\

\begin{figure*}
    \centering
    \begin{subfigure}[t]{0.38\textwidth} 
        \centering
        \includegraphics[width=0.75\textwidth]{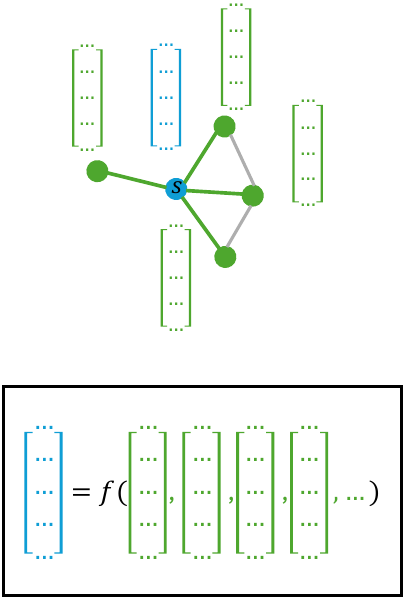}
        \caption{} 
        \label{fig:gnn}
    \end{subfigure}
    \hfill
    \begin{subfigure}[t]{0.61\textwidth} 
        \centering
        \includegraphics[width=0.7\textwidth]{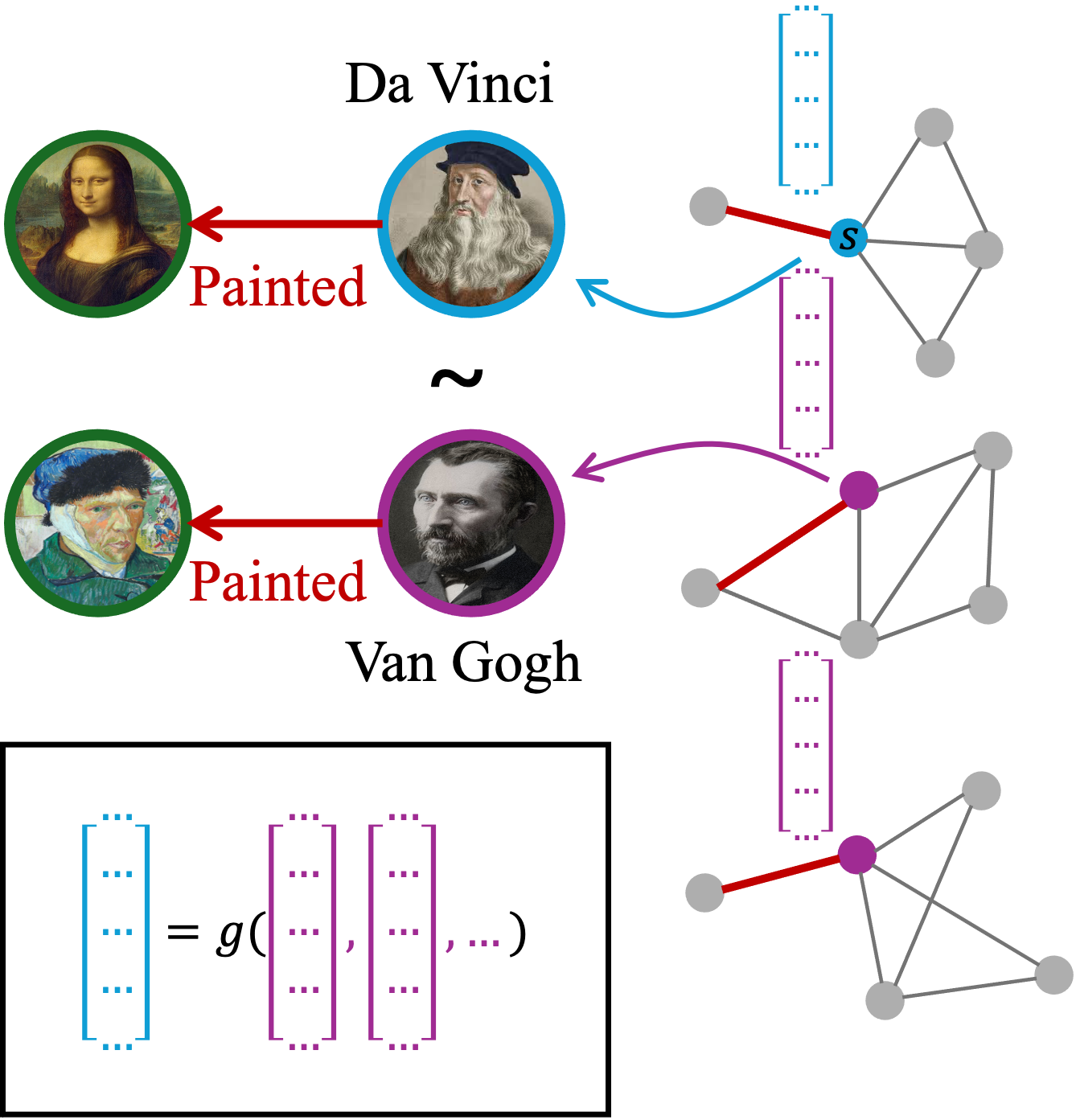}
        \caption{} 
        \label{fig:similar}
    \end{subfigure}  

    \caption{Local vs. Relation-Based Entity Similarity Enhancements.(a) A traditional neighborhood-based approach (function $f$ relies on direct connections. (b)The proposed enhancement function $g$ leverages relation-based similarity (e.g., “painted”), linking entities like Da Vinci and Van Gogh beyond their immediate neighbors.}
    \label{fig:similarity}
    
\end{figure*}

\noindent\textbf{Relation-Based Entity Similarity}. This component identifies entities as similar when they engage in identical interaction types with different objects. In other words, two entities are considered similar if they participate as subjects in events involving the same relation type. For example, consider the events \textit{(da Vinci, painted, Mona Lisa, 1503)} and \textit{(Van Gogh, painted, Self-Portrait with Bandaged Ear, 1889)}; while \textit{da Vinci} and \textit{Van Gogh} might not be directly connected, they are deemed similar because they both have \textit{``painted''} connections. This shared interaction pattern implies they occupy analogous positions in the relational structure of the graph. From a reasoning perspective, identifying such relational similarities allows the model to generalize learned knowledge to unseen or rare entities. If a model understands that \textit{da Vinci} and\textit{Van Gogh} both \textit{``painted''}  something, it may infer similar properties or roles for them. The set of similar entities at time $t$ with respect to relation $r$ is defined as:
\begin{equation}
S_t(r) = \{s_i | (s_i, r, o_i, t_i) \in G, t_i < t \}
\end{equation}

\noindent\textbf{Enhancement Function}. The enhanced representation of a subject entity \( s \) is computed within the context of the query \cite{mirtaheri2024tackling}. More specifically, for a given query \( (s, r, ?, t) \), the enhancement function \( g(s, r, t) \) is computed with respect to relation \( r \) and time \( t \), rather than using the generic term \( g(s) \) in Equation \ref{eq:enhancement_layer}. The function aggregates representations of entities that with similar interactions:  

\begin{equation}
g(s, r, t) = \frac{\sum_{s_i \in S_t(r)} w_i e_{s_i}}{\sum w_i}, \quad w_i = \frac{1}{1 + \exp (\mu(t - t_i))}
\end{equation}  

Here, \( e_{s_i} \) represents the embedding of entity \( s_i \), where \( (s_i, r, o_i, t_i) \in S_t(r) \), and \( w_i \) is a weight that accounts for the recency of the interaction \( (s_i, r, o_i, t_i) \). To maintain computational efficiency, the implementation processes only a subset of \( S_t(r) \), prioritizing recent interactions and their associated entities.

\subsection{Weighted Frequency-Based Sampling}

Standard training approaches for TKG completion methods typically suffer from an unbalanced emphasis on frequently occurring entities—those that participate in many interactions across the graph.
The reason is that the standard training process   disproportionately tunes the model towards entities that appear frequently in the training dataset.
This skewed focus leads to models that perform well on common entities but struggle with rare ones. To address this imbalance, we develop a weighted frequency-based sampling framework that systematically prioritizes training examples involving less common entities. Our approach selectively samples quadruples with probability inversely proportional to the frequency of their constituent entities, thereby increasing representation of long-tail entities during the training process. Our sampling algorithm operates in two phases:
\begin{enumerate}
    \item We first perform weighted sampling to select $\alpha$ percent of the training data with replacement. For each quadruple $(s, r, o, t)$, we assign a sampling weight determined by $\psi\left(\frac{1}{\text{freq}(s)}, \frac{1}{\text{freq}(o)}\right)$, where $\psi$ can be instantiated as a minimum, maximum, or mean function. 
This weighting ensures that quadruples involving rare entities are more likely to be sampled, thereby increasing their presence in the training set and giving the model more opportunities to learn from long-tail patterns.
    
    \item We then apply uniform sampling to select the remaining $(1-\alpha)$ percent of training examples with replacement. This balanced approach prevents the training process from focusing exclusively on rare entities, which would risk overfitting to infrequent patterns.
\end{enumerate}

In our incremental training framework, entity frequencies are calculated dynamically based on quadruples observed up to the current training step, rather than using global statistics from the entire dataset. This design choice ensures that the sampling strategy adapts to evolving entity distributions across temporal snapshots.  This incremental calculation enables the sampling strategy to adapt as the entity distribution evolves over time.  The two-phase sampling process also ensures that the model neither neglects rare entities nor overfits to them, promoting more robust and generalizable TKG completion.

\section{Experimental Setup}\label{sec:exp_setup}
We describe the datasets, evaluation setup, results, and subsequent analysis for our experiments.




\begin{table}[t]
\centering
\caption{Dataset Statistics}
    \begin{tabular}{l | c c c c c }
    \toprule
        Dataset & $|\mathcal{E}|$ &  $|\mathcal{R}|$ & \#Snapshots & $|\mathcal{Q}|_{G_1}$ & Avg $\mathcal{|Q|}_{G_{i>1}}$\\
        \midrule

        \icewsd & 7128 & 230 &	33	& ~28k/3.7k/4k &	~1k/0.3k/0.3k \\
        \icewsw & 23039	& 230 & 16	& ~244k/45k/43k & 	~8K/1k/1k \\
\bottomrule
\end{tabular}
    
\label{tab:datastats}

\end{table}

\begin{table*}[t]
\centering
\caption{Total Link prediction performance comparison. Hit@10, Hit@3, Hit@1 and MRR reported for different models incrementally trained using two benchmarks: \icewsw and \icewsd. Performance is evaluated at the final training time step over the last test dataset(Current) and across all prior test datasets (Average). FT: Fine-Tuning, EWC: Elastic Weight Consolidation, ER: Experience Replay. The first line is the model trained only on the first snapshot.}
\fontsize{8pt}{8pt}\selectfont
\setlength\tabcolsep{1.5pt}
    \begin{tabular}{l l | c c c c c c c c | c c c c c c c c }
        \toprule
         \multicolumn{2}{c}{} & \multicolumn{8}{c}{\icewsw} & \multicolumn{8}{c}{\icewsd} \\
        \cmidrule(lr){3-10}
        \cmidrule(lr){11-18}
        \multicolumn{2}{c}{} & \multicolumn{4}{c}{Current} & \multicolumn{4}{c}{Average} & \multicolumn{4}{c}{Current} & \multicolumn{4}{c}{Average} \\
        \cmidrule(lr){3-6}
        \cmidrule(lr){7-10}
        \cmidrule(lr){11-14}
        \cmidrule(lr){15-18}
        \multicolumn{2}{c}{Model} & H@10 &  H@3 & H@1 & MRR & H@10 & H@3 & H@1& MRR & H@10 & H@3 & H@1 & MRR & H@10 & H@3 & H@1 & MRR \\
        \midrule

Titer  &  &0.443 &0.317 &0.206 &0.287 &0.451 &0.336 &0.224 &0.303 &0.571 &0.466 &0.364 &0.436 &0.555 &0.428 &0.286 &0.380 \\ 
 & + FT & 0.444 &0.324 &0.214 &0.295 &0.464 &0.350 &0.229 &0.312 &0.582 &0.494 &0.393 &0.464 &0.572 &0.456 &0.326 &0.413 \\ 
\cmidrule{2-18}
& + ER & 0.487 &0.353 &0.234 &0.319 &0.470 &0.349 &0.231 &0.314  &0.593 &0.477 &0.379 &0.452 &0.572 &0.443 &0.304 &0.396 \\
& + EWC & 0.484 &0.364 &0.239 &0.326 &0.476 &0.357 &0.234 &0.319  &0.588 &0.486 &0.395 &0.463 &0.574 &0.451 &0.311 &0.403  \\
\cmidrule{2-18}
& + Ours & \textbf{0.496} &\textbf{0.368} &\textbf{0.242} &\textbf{0.330} &\textbf{0.489} &\textbf{0.362} &\textbf{0.236} &\textbf{0.323} &\textbf{0.610} &\textbf{0.517} &\textbf{0.407} &\textbf{0.482} &\textbf{0.581} &\textbf{0.469} &\textbf{0.334} &\textbf{0.421} \\

\bottomrule
\end{tabular}

\label{tab:results-total}
\end{table*}


\subsection{Dataset Construction}\label{sec:dataset}

We use the Integrated Crisis Early Warning System (ICEWS) dataset, which records interactions among geopolitical actors with daily event timestamps. For evaluation benchmarks, we select a specific interval from the ICEWS dataset. We then generate temporal snapshots by partitioning the data into sets of quadruples with unique timestamps. The initial snapshot contains 50\% of the selected interval, and the remaining data is segmented into non-overlapping periods of size \( w \). Each snapshot is further divided into training, validation, and test sets based on their timestamps. In line with the extrapolation setup, the training set timestamps are earlier than those in the validation set, which are in turn earlier than the test set timestamps, as depicted in Figure \ref{fig:cl-ext}. For both ICEWS14 and ICEWS18, the first seven months are designated as the initial snapshot, while subsequent snapshots each span a seven-day period. Despite both benchmarks originating from the ICEWS dataset, ICEWS14 and ICEWS18 exhibit significant differences in terms of density and distribution, as detailed in Table \ref{tab:datastats}, which presents the statistics of both datasets.




\subsection{Evaluation Setup}\label{sec:eval}
Our primary objective is to assess our model's performance under both continual and inductive learning scenarios. To our knowledge, this is the first study that investigates inductive and continual learning for temporal knowledge graphs in an extrapolation setup. Our approach aligns closely with prior works ~\citep{wu2021tie,mirtaheri2023history,cui2023lifelong} that explore continual learning for node classification in heterogeneous graphs and semantic knowledge graph completion.

We partition the quadruples of the Temporal Knowledge Graph (TKG) into sequential temporal snapshots, as detailed in Section \ref{sec:dataset}. Each snapshot functions as a distinct task for incremental training, with subsequent snapshots introducing new facts and potentially new entity links.

\textbf{Incremental Training Procedure}: Training commences on the model \(\mathcal{M}\) using \(\tr_1\), with \(\val_1\) employed for hyper-parameter tuning. At each time step \(t\), the model \(\mathcal{M}_t\) with parameter set \(\theta_t\) is initialized with parameters from the preceding time step \(\theta_{t-1}\). Subsequently, \(\mathcal{M}_t\)'s parameters are updated by training on \(\tr_t\). As illustrated in Figure \ref{fig:cl-ext}, in the extrapolation setup, the training, validation, and testing datasets are constructed to ensure no temporal overlap. Specifically, event timestamps in the training data precede those in both the validation and test datasets. This design ensures the model is trained on historical data and evaluated on future quadruples with unseen timestamps. However, for incremental learning, this approach may omit data segments containing pivotal information. Thus, at training step \(t\), we maintain two checkpoints: one post-training on \(\tr_t\) for evaluation and another post-training on both the validation and test sets for a few epochs before proceeding to \(\tr_{t+1}\). This incremental training can be a straightforward fine-tuning or can incorporate advanced incremental training strategies. For our base model, we employ TiTer ~\citep{sun2021timetraveler}, which is the state-of-the-art in inductive temporal knowledge graph completion. TiTer leverages reinforcement learning to facilitate path-based reasoning.

\begin{table*}[t]
\centering  
\caption{Inductive Link prediction performance comparison. Hit@10, Hit@3, Hit@1 and MRR reported for different models incrementally trained using \icewsw and \icewsd. Performance is evaluated at the first inductive test set (First) and the union of all inductive test sets at the last training time step (Average). FT: finetuning, EWC: Elastic Weight Consolidation, ER: Experience Replay.}
\fontsize{8pt}{8pt}\selectfont
\setlength\tabcolsep{1.5pt}
    \begin{tabular}{l l  c c c c c c c c | c c c c c c c c }
        \toprule
         \multicolumn{2}{c}{} & \multicolumn{8}{c}{\icewsw} & \multicolumn{8}{c}{\icewsd} \\
        \cmidrule(lr){3-10}
        \cmidrule(lr){11-18}
        \multicolumn{2}{c}{} & \multicolumn{4}{c}{First} & \multicolumn{4}{c}{Average} & \multicolumn{4}{c}{First} & \multicolumn{4}{c}{Average} \\
        \cmidrule(lr){3-6}
        \cmidrule(lr){7-10}
        \cmidrule(lr){11-14}
        \cmidrule(lr){15-18}
        \multicolumn{2}{c}{Model} & H@10 &  H@3 & H@1 & MRR & H@10 & H@3 & H@1& MRR & H@10 & H@3 & H@1 & MRR & H@10 & H@3 & H@1 & MRR \\
        \midrule

         Titer & + FT & 0.448 &0.334 &0.219 &0.298 &0.091 &\textbf{0.073} &0.048 &0.064 & 0.547 &0.421 &0.285 &0.376 &0.146 &0.126 &0.097 &\underline{0.115} \\
         \cmidrule{2-18}
        & + ER &  & &  & &0.090 &0.069 &0.045 &0.061 &  & & & &\textbf{0.147} &0.118 &0.081 &0.103 \\
        
        & + EWC &  & & & &0.089 &0.070 &0.044 &0.060 &  & & & &0.144 &0.109 &0.074 &0.098 \\
        \cmidrule{2-18}
        & + Ours &\textbf{0.458} &\textbf{0.342} &\textbf{0.223} &\textbf{0.305} &\textbf{0.092} &0.071 &\textbf{0.054} &\textbf{0.066} & \textbf{0.564} &\textbf{0.439} &\textbf{0.295} &\textbf{0.388} &0.146 &\textbf{0.128} &\textbf{0.098} &\textbf{0.115} \\

\bottomrule
\end{tabular}

\label{tab:results-inductive}
\end{table*}

\begin{table*}[t]
\centering
\caption{Ablation study on different model components. Compares Hit@10, Hit@3, Hit@1 and MRR for the base model, Titer, augmented with Fine-Tuning (FT) against our framework with isolated components (Weighted Sampling and Enhancement Layer) and their combined effect.}
\fontsize{7.5pt}{9pt}\selectfont
\setlength\tabcolsep{0.75pt}
    \begin{tabular}{l l | c c c c c c c c | c c c c c c c c }
        \toprule
         \multicolumn{2}{c}{} & \multicolumn{8}{c}{\icewsw} & \multicolumn{8}{c}{\icewsd} \\
        \cmidrule(lr){3-10}
        \cmidrule(lr){11-18}
        \multicolumn{2}{c}{} & \multicolumn{4}{c}{Current} & \multicolumn{4}{c}{Average} & \multicolumn{4}{c}{Current} & \multicolumn{4}{c}{Average} \\
        \cmidrule(lr){3-6}
        \cmidrule(lr){7-10}
        \cmidrule(lr){11-14}
        \cmidrule(lr){15-18}
        & Model & H@10 &  H@3 & H@1 & MRR & H@10 & H@3 & H@1& MRR & H@10 & H@3 & H@1 & MRR & H@10 & H@3 & H@1 & MRR \\
        \midrule

Titer & + FT & 0.444 &0.324 &0.214 &0.295 &0.464 &0.350 &0.229 &0.312 &0.582 &0.494 &0.393 &0.464 &0.572 &0.456 &0.326 &0.413 \\ 
\cmidrule{2-18}
& + Ours (Weighted Sampling) &0.482 &0.366 &0.237 &0.325 &0.477 &0.359 &0.236 &0.320 &0.621 &0.489 &0.393 &0.467 &0.579 &0.459 &0.321 &0.412  \\
  & + Ours (Enhancement Layer) & 0.490 &0.367 &0.246 &0.330 &0.484 &0.365 &0.239 &0.325 &0.579 &0.483 &0.393 &0.459 &0.583 &0.468 &0.343 &0.427 \\
 &+ Ours (Full) & 0.496 &0.368 &0.242 &0.330 &0.489 &0.362 &0.236 &0.323 &0.610 &0.517 &0.407 &0.482 &0.581 &0.469 &0.334 &0.421 \\

\bottomrule
\end{tabular}
    
\label{tab:results-ablation}

\end{table*}


\textbf{Baseonines for Comparison}. In addition to the naive fine-tuning (\textsc{FT}) approach, we implement several variants of the base model, each augmented with distinct \textit{Continual Learning (CL)} strategies aimed at mitigating catastrophic forgetting and enhancing temporal generalization.
These strategies include   \textbf{Elastic Weight Consolidation (EWC)} method~\citep{kirkpatrick2017overcoming}, to represent  regularization-based techniques
We also incorporate an \textbf{Experience Replay (ER)} ~\citep{rolnick2019experience}, which stores a subset of previously observed data and reuses it during training to reinforce earlier knowledge.
Together, these CL methods provide a comparative framework to evaluate the efficacy of memory- and regularization-based approaches in the context of temporal knowledge graph learning.

\textbf{Evaluation Metrics}. Similar to other KG completion studies ~\citep{liang2022reasoning}, we evaluate the models using Mean Reciprocated Rank (MRR) and Hit@k metrics for $k=1, 3, 10$. Following ~\citep{mirtaheri2023history}, and in order to assess the ability of the model in alleviating the forgetting problem, at the current snapshot $t$, we report the average model performance over all the test data from previous snapshots, as well as the current test data.

\textbf{Implementation Detail}. We adopted the hyperparameters of the base model, Titer, as outlined in their paper and the accompanying code repository. For the enhancement layer and weighted sampling, we performed an extensive grid search. The search space encompassed: $\lambda \in [ 0.3, 0.5, 0.7 ]$, $\mu \in [0.1, 0.3, 0.5]$, the number of similar entities within $[10, 15, 20, 25]$, and the weighted sampling fraction $\alpha \in [0, 0.1, 0.2, 0.5, 0.8, 1]$. Model selection was based on achieving the highest average incremental performance on the validation set.




\section{Experiments} \label{sec:exp}
Our experiments focus on entity prediction, commonly referred to as knowledge graph completion. Through these experiments, we aim to address the following research questions:

\begin{itemize}
    \item \textbf{Q1 (Section \ref{sec:q1})}: Can our framework effectively generalize to new facts while preserving previously acquired knowledge?
    \item \textbf{Q2 (Section \ref{sec:q2})}: How well does our approach generalize to links involving new entities and entities with sparse neighborhoods?
    \item \textbf{Q3 (Section \ref{sec:q3})}: What is the contribution of each component of our framework to specific challenges, such as incremental learning and handling long-tail entities?
\end{itemize}

We provide experimental results and subsequent analysis to address the above questions. 
 
\subsection{Overall Performance (Q1)}\label{sec:q1}
Our evaluation centers on assessing the link prediction performance of various models trained under an incremental learning paradigm, using two widely adopted temporal knowledge graph benchmarks: \icewsw and \icewsd. The key findings are presented in Table~\ref{tab:results-total}.
To establish a baseline, we first examine the performance of the original Titer model when trained solely on the initial snapshot. As shown in the first row of the table, Titer exhibits notably poor performance on both the most recent snapshot and the average across all test sets. This underscores the model’s limited ability to generalize across time, indicating a substantial challenge in retaining or transferring knowledge effectively to future timesteps.
For all other model variants, Table~\ref{tab:results-total} reports the performance of the final model after incremental training, denoted as $\calM_T$, evaluated both on the final snapshot and averaged across all timesteps. Among these, simple fine-tuning (FT) of the original Titer model leads to only marginal improvements, suggesting that naively updating parameters over time is insufficient for robust temporal generalization.
In contrast, ER and EWC offer improvements by introducing mechanisms for preserving past knowledge. ER benefits from replaying previous data samples, while EWC constrains updates to parameters critical for past tasks—both strategies mitigate catastrophic forgetting to varying degrees.
Most notably, integrating our proposed framework into the base model yields the best performance across all configurations. Our method demonstrates significant gains, particularly in Mean Reciprocal Rank (MRR) and Hit@10, the standard metrics for link prediction. Specifically, we observe a 10\% relative improvement in MRR on \icewsd and a 15\% improvement on \icewsw, highlighting the effectiveness of our framework in enhancing link prediction performance within an incremental training setting.

\begin{figure*}
    \centering
    \begin{subfigure}[t]{0.6\textwidth}
        \centering
        \includegraphics[trim={0 0 0 15pt},clip, width=1\linewidth]{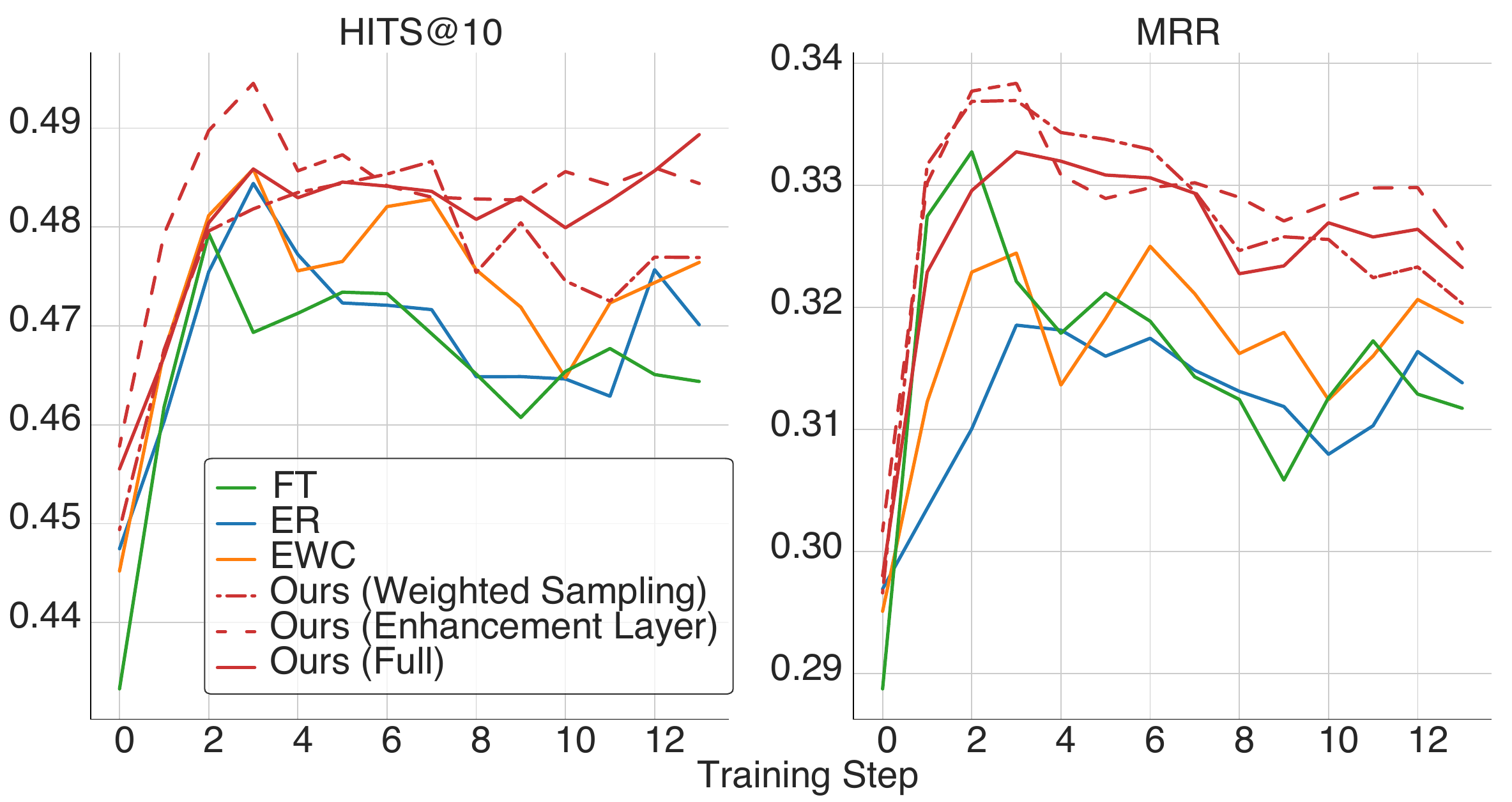}
        \caption{Continual Learning Performance on \icewsw}
        \label{fig:continual}
    \end{subfigure}%
    ~ 
    \begin{subfigure}[t]{0.4\textwidth}
        \centering
        \includegraphics[trim={0 20pt 0 0},clip,width=1\linewidth]{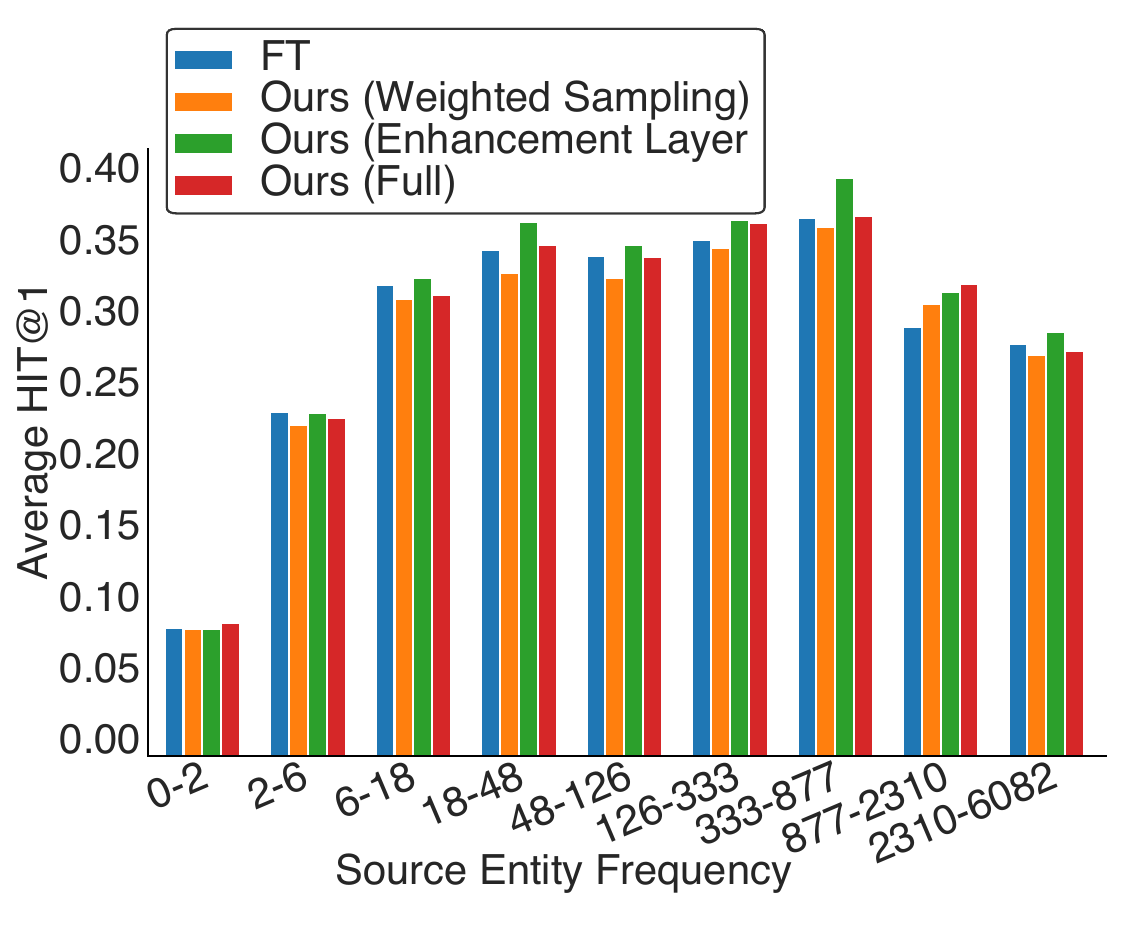}
        \caption{Hit@1 Performance over the union of test datasets grouped by source entity frequency}
        \label{fig:longtail}
    \end{subfigure}
    
    \caption{Analysis of model components for catastrophic forgetting and long-tail entities.}
    \label{fig:main}
\end{figure*}

\subsection{Inductive Link Prediction (Q2)}\label{sec:q2}
The inductive link prediction performance of various models, incrementally trained using the benchmarks \icewsw and \icewsd, is presented in Table \ref{tab:results-inductive}. Unlike transductive settings, where all entities are observed during training, the inductive setting requires the model to predict links involving entities not present in any prior training data.  This evaluation assesses the model's ability to predict links involving entities that were not observed during training. In our incremental framework, an entity $s$ at training step $t$ is considered unseen if it hasn't appeared in graph in previous steps and also the current training data, i.e., $s \notin \bigcup_{j=1}^{t-1} \mathcal{Q}_j \cup \tr_t$.

 We report the inductive link prediction performance for two key model checkpoints: $\mathcal{M}_1$ and $\mathcal{M}_T$. The model $\mathcal{M}_1$ is trained exclusively on the initial snapshot of the dataset and evaluated on the corresponding test set $\te_1$. This result is denoted as ``First'' in Table~\ref{tab:results-inductive}. The initial snapshot holds particular importance in our study as it represents the largest subset of the data, containing the highest number of quadruples among all temporal segments. As such, it provides a strong foundation for assessing early-stage model performance in the inductive setting.

For the first snapshot, we observe that the results for FT, EWC, and ER  are identical. This outcome is expected, as all three methods begin from the same model initialization and are trained on the identical initial dataset using the same underlying architecture. Since no additional mechanisms—such as regularization (in the case of EWC) or memory buffers (in the case of ER)—are activated at this early stage, the behavior of these models is effectively indistinguishable from standard training. As a result, to avoid redundancy, we omit these repeated results from Table~\ref{tab:results-inductive}.
In contrast, our proposed framework introduces its core components—weighted sampling and the enhancement layer—from the very first training step. Unlike FT, EWC, and ER, our approach deviates from baseline training immediately for the first snapshot. This distinction highlights the early-stage benefits of our framework and its potential to improve inductive generalization from the outset.

When integrated with the base model, Titer, our method yielded substantial improvements across all metrics for the first snapshot. It also surpassed other baselines in the incremental task for \icewsw, particularly in MRR and Hit@1. However, the task of incremental inductive link prediction proved to be more challenging, evident from the significant performance drop across all models. This complexity likely stems from the emergence of links in certain snapshots that are inherently more difficult to predict, leading to a pronounced decline in the performance metrics for all models.

\begin{table*}[h]
\centering
\fontsize{8pt}{8pt}\selectfont
\caption{Comparative Analysis of Time and Memory Complexities for various models: includes the enhancement layer, weighted sampling, and the full model. Also reports per epoch and total runtime for the naive Titer and Titer integrated with our models, applied to the ICEWS14 dataset}
\begin{tabular}{l|c c c c c}
\toprule
\textbf{Methodology}        & \textbf{Time Complexity} & \textbf{Memory Complexity} & \textbf{Per Epoch Runtime}& \textbf{Total Runtime} \\ 
\midrule
Naive Titer                 & -                              & -                                   & $\sim$9s                   & $\sim$90m              \\ 
Enhancement Layer           & $\mathcal{O}(bnd)$             & $\mathcal{O}(bnd)$                  & $\sim$12s                  & $\sim$125m             \\ 
Weighted Sampling           & $\mathcal{O}(n)$               & $\mathcal{O}(\|\mathcal{E}\|)$      & $\sim$9s                   & $\sim$110m             \\ 
Full                        & $\mathcal{O}(bnd)$             & $\mathcal{O}(\|\mathcal{E}\| + bnd)$& $\sim$12s                  & $\sim$125m             \\ 
\bottomrule
\end{tabular}
\label{tab:compute}
\end{table*}

\subsection{Ablation Study (Q3)}\label{sec:q3}

To elucidate the individual contributions of the components within our proposed framework, we conducted an ablation study. The results are presented in Table \ref{tab:results-ablation}. We evaluated the base model, Titer, augmented with Fine-Tuning (FT) as a baseline. Subsequently, we assessed the performance of our framework with individual components: Weighted Sampling, Enhancement Layer, and the full combination of both. For the full model the hyper parameters are selected that achieve the best overall link prediction performance. For the isolated components, the hyper parameters were kept identical to the parts of the full model.

The results indicate that each component of our framework contributes to the overall performance improvement. Specifically, the Enhancement Layer and Weighted Sampling individually boost the model's performance, with the full combination of both components achieving the best results across all metrics. This ablation study underscores the synergistic effect of our proposed components in enhancing link prediction performance, especially in the context of incremental training.

To further analyze the contributions of each component in incremental learning, we assess how effectively each model mitigates catastrophic forgetting. Specifically, at time  $t$, we compute the average performance of $\mathcal{M}_t$ over the current test set and all preceding ones, i.e., $\te_1, \te_2, \dots, \te_t$. We define the performance at time $t$ as $P_t = \frac{1}{t} \sum_{j=1}^t p_{t, j}$, where $p_{t, j}$ represents the performance of $\mathcal{M}_t$ on $\te_j$, measured by metrics such as MRR or Hit@10. As illustrated in Figure \ref{fig:continual}, our model's components, both individually and in combination, excel over other baselines in mitigating catastrophic forgetting, evident from the reduced performance decline over time. Notably, the enhancement layer demonstrates pronounced efficacy in this regard. While EWC surpasses other baselines, Experience Replay (ER) and Fine-Tuning (FT) exhibit marginal performance variations.

Furthermore, we examine the efficacy of each component in catering to long-tail entities. Figure \ref{fig:longtail} delineates the performance of each model on the union of test sets $\te_1, \dots, \te_T$. Test set quadruples are aggregated based on the incremental frequency of their source entity. This incremental frequency, distinct from the overall frequency, corresponds to the entity's occurrence rate in the graph when the specific query quadruple was observed. The findings reveal that for entities with frequencies below 18, none of the models show a significant difference in the Hit@1 score. However, for frequencies exceeding 18, our model consistently surpasses the fine-tuned variant. The enhancement layer, in particular, significantly outperforms the fine-tuned model, especially for frequencies within the [18-48] range and beyond.

    
    




\section{Complexity and Runtime Analysis}
The storage complexity of the enhancement layer is $\mathcal{O}(|\mathcal{B}|nd)$ where $\mathcal{B}$ is the batch and $n$ is the maximum number of similar entities, and the time complexity of the enhancement layer is $\mathcal{O}(|\mathcal{B}|nd)$ for one forward pass, where $\mathcal{B}$ is the batch, $n$ is the maximum number of similar entities per entity and $d$ is the entity embedding dimensions. At each forward pass, for each quadruple in the batch, the algorithm has access to $n$ other similar entities and retrieves and keep their embeddings in the memory. Each forward pass of the enhancement layer involves elementwise multiplication of the retrieved entities with the exponential coefficients and then a temporal averaging which takes $\mathcal{O}(|\mathcal{B}|nd)$. Table \ref{tab:compute} reports the memory and time complexity of each step of our approach both theoretically and in dataset ICEWS14.

\section{Conclusion} \label{sec:conclusion}

In this work, we developed an incremental training framework for temporal knowledge graphs by incorporating a model-agnostic enhancement layer and a weighted sampling strategy. When augmented to GNN-based TKG completion methods, our approach yielded improvements in overall link prediction performance. Compared to models trained without incremental adaptation or other continual learning baselines, our method demonstrated better performance in mitigating catastrophic forgetting, with the enhancement layer being particularly influential in this aspect. Our framework also exhibited improved results in inductive link prediction and effectively addressed long-tail entities, especially those with moderately populated neighborhoods. While our approach marks a step forward in TKG completion, there remain areas for improvement, particularly for entities with extremely sparse neighborhoods. Future research directions could include the integration of richer features and continuous extraction of facts from LLMs to further refine TKG representations.



\bibliography{paper}
\bibliographystyle{ACM-Reference-Format}


\end{document}